\documentclass{article}
\usepackage[final,nonatbib]{neurips_2018}

\usepackage[utf8]{inputenc} 
\usepackage[T1]{fontenc}    
\usepackage{hyperref}       
\usepackage{url}            
\usepackage{booktabs}       
\usepackage{amsfonts}       
\usepackage{nicefrac}       
\usepackage{microtype}      
\usepackage{amsmath,amssymb}
\usepackage{algorithmic}
\usepackage{subcaption}
\usepackage{graphicx}

\usepackage{multirow}
\usepackage[boxruled,linesnumbered]{algorithm2e}

\newcommand{\K}{{\mathcal K}}
\newcommand{\MB}[1]{\mathbf{#1}}
\newcommand{\MBB}[1]{\mathbb{#1}}
\newcommand{\MC}[1]{\mathcal{#1}}

\newcommand{\smalleqb}[1]
{
	\begingroup
	\makeatletter
	\def
	\f@size{1}
	{#1}
    \endgroup
}

\newcommand{\nobj}{\MC{J}}

\newcommand{\nZ}{{\MB{Z}}}
\newcommand{\nZC}{\mathcal{Z}}
\newcommand{\ny}{{\vec{x}}}
\newcommand{\nY}{\mathbf{X}}
\newcommand{\nYC}{\mathcal{X}}
\newcommand{\nxs}{\gamma}
\newcommand{\nx}{{\vec{\gamma}}}
\newcommand{\nX}{\mathbf{\Gamma}}

\newcommand{\nA}{\mathbf{U}}
\newcommand{\na}{{\vec{u}}}
\newcommand{\nas}{u}
\newcommand{\nT}{T_0}

\newcommand{\nALF}{\mathbf{B}}
\newcommand{\nAlf}{\vec{\beta}}
\newcommand{\nalf}{{\beta}}

\newcommand{\phiba}{\Phi_{\MB{\nALF}}(\nA)}

\SetKwInput{KwInput}{Input}
\SetKwInput{KwParameters}{Parameters}
\SetKwInput{KwOutput}{Output}
\SetKwInput{KwTask}{Task}
\SetKwInput{KwInit}{Initialization}
\SetKwInput{KwProc}{Loop}
\SetKwInput{KwStepk}{Step K}
\SetKwInput{KwPre}{Pre-calculations}
\SetKwInput{Kwinit}{Initialization}

\title{Multiple-Kernel Dictionary Learning for\\ Reconstruction and Clustering of Unseen Multivariate Time-series}

\author{%
	Babak Hosseini
	\thanks{
		Preprint of the publication~\cite{hosseini2019LMMK}, as provided by the authors.
		The final publication is available at \url{https://www.elen.ucl.ac.be/esann/index.php?pg=proceedings}
	} \\
	CITEC cluster of excellence\\
	Bielefeld University, Germany\\
	\texttt{bhosseini@techfak.uni-bielefeld.de} \\
	\And
	Barbara Hammer\\
	CITEC cluster of excellence\\
	Bielefeld University, Germany\\
	\texttt{bhammer@techfak.uni-bielefeld.de} \\
}

\begin{document}
\maketitle

\begin{abstract}
There exist many approaches for description and recognition of unseen classes in datasets. 
Nevertheless, it becomes a challenging problem when we deal with multivariate time-series (MTS) (e.g., motion data), 
where we cannot apply the vectorial algorithms directly to the inputs.  	
In this work, we propose a novel multiple-kernel dictionary learning (MKD) which
learns semantic attributes based on specific combinations of MTS dimensions in the feature space.
Hence, MKD can fully/partially reconstructs the unseen classes based on the training data (seen classes).  
Furthermore, we obtain sparse encodings for unseen classes based on the learned MKD attributes, 
and upon which we propose a simple but effective incremental clustering algorithm to categorize the unseen MTS classes in an unsupervised way.
According to the empirical evaluation of our MKD framework on real benchmarks, it provides an interpretable reconstruction of unseen MTS data as well as a high performance regarding their online clustering.
\end{abstract}

\section{Introduction}
Zero-shot learning is the problem of recognizing novel categories of data when no prior information is available during the training phase~\cite{alabdulmohsin2016attribute,lampert2009learning,socher2013zero}. 
One practical approach to such transfer learning is the incorporation of semantic attributes as descriptive features to 
map the input data to an intermediate semantic space, 
which can discriminate between different unseen categories \cite{lampert2009learning,socher2013zero}. 
Another concern in this area of research is the partial/complete reconstruction of the unseen classes based on their relation to the learned semantic attributes or the training data \cite{peng2018joint,qiu2011sparse}. 

\begin{figure}[!t]
	\centering		
	\includegraphics[width=1\linewidth]{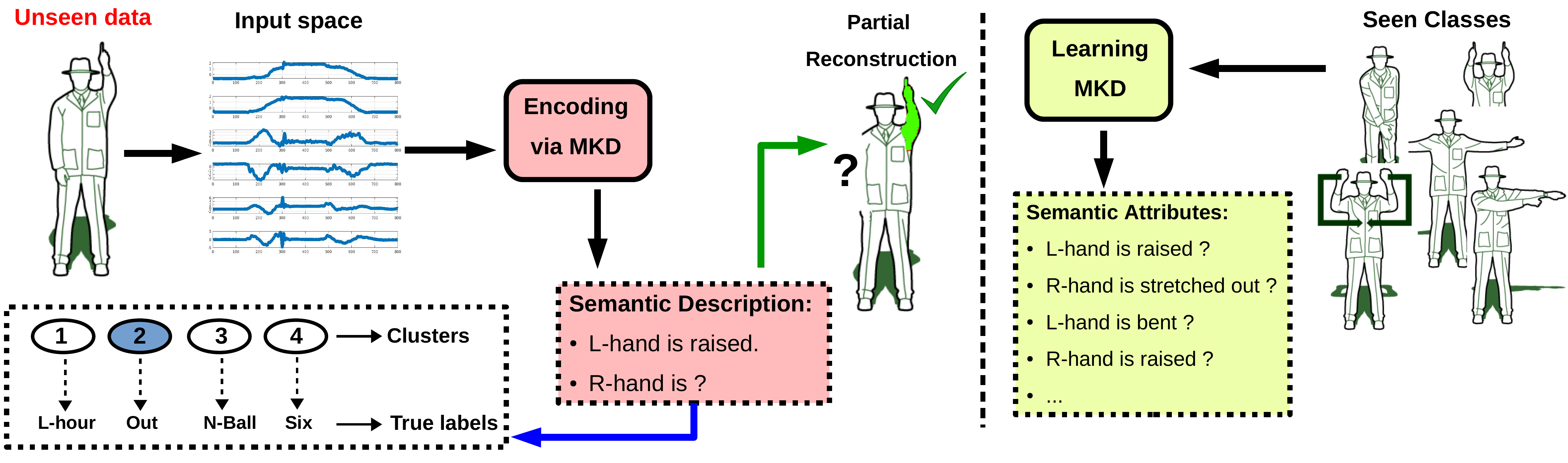}
	\caption{General overview of our framework. The dictionary (MKD) learns the semantic attributes based on the seen classes. These attributes are used for a semantic description of the data from the unseen classes, which leads to categorizing and partial reconstruction of the data.}
	\label{fig:alg}
\end{figure}

An important application of zero-shot learning is multivariate time-series (MTS) in the general meaning such as audio data and human motions \cite{cheng2013nuactiv,lu2016self} with a considerable number of unknown classes. 
Different from images and video, MTS do not possess any general spatial dependency between its dimensions. 
Nevertheless, it is usually expected to find semantic attributes shared between different classes of an MTS dataset. 
As an example of MTS data, consider the Cricket Umpire signal \textit{Out} in Fig.~\ref{fig:alg} which can be described as the \textit{left hand is raised} while \textit{the right hand is down}. 
Such encoding provides us with a semantic understanding of the data without having any prior knowledge about its class label. 
We can also consider such descriptions as semantic attributes in order to distinguish the unknown MTS data samples into distinct categories that reflect their unknown labels. 
Although the semantic descriptions are class specific, we can share the individual attributes among classes which have between-class partial similarities.

Sparse coding (SRC) is the idea of constructing an input data using weighted combinations (\textit{sparse codes}) of sparse selected entries from a set of learned bases (\textit{dictionary}). 
Such sparse representations can capture essential intrinsic characteristics of a dataset \cite{rubinstein2008efficient}.
Furthermore, via assuming an implicit mapping of the data to a high-dimensional feature space, it is possible to formulate SRC using the kernel representation of the data \cite{jian2011design} to  
model also nonlinear data structures. 
Consequently, a subset of the existing research has benefited from SRC methods in designing more effective attributes for dealing with unseen classes of data; however, these efforts are mainly limited to the image (spatial) and video (spatiotemporal) datasets \cite{qiu2011sparse,zhang2015zero}

Despite the current achievements in learning unseen MTS data,  
either the existing methods are depended on having prior information about the novel classes (e.g., samples/labels) \cite{lu2016self}, 
or they cannot interpret the unseen data based on their learned attributes.
Furthermore, to our knowledge, there is no research reported on the partial/complete reconstruction of unseen classes for MTS data in general (e.g., recorded motion signals).

To address the above concerns, we provide the following contributions:
\\	\textbf{1-} We design a novel dictionary structure which learns attributes that can represent MTS based on the dimension level.
\\	\textbf{2-} We propose an unsupervised kernel-based SRC method for partial reconstruction of unseen MTS data in the feature space along with their interpretable encoding. 
\\	\textbf{3-} We design an incremental clustering based on the sparse encodings of the unseen data which gradually creates a clustering dendrogram of the unseen classes.

After formulating the problem in Sec.~\ref{sec:prob},
we introduce and explain our proposed framework in Sec.~\ref{sec_exp}, and we evaluate it in Sec. \ref{sec_exp} followed by the conclusion section.
\section{Problem Statement}\label{sec:prob}
Presenting a multivariate time-series in the vectorial space, 
$\nY_i=[\ny_i(1)\dots\ny_i(T)] \in \mathbb{R}^{f\times T}$ denotes sequence $i$,
where $(f,T)$ represents the number of dimensions and the sequence lengths respectively.
The training set $\nYC=\{\nY_i\}_{i=1}^N$ belongs to $c$ distinct data classes with the label set $l=\{1,\cdots,c\}$. 
Accordingly, the set of unseen MTS $\nZC$ belongs to the label set $q$,
such that $q \cap l= \emptyset$.
Based on the above description, we are interested in:
	\textbf{1-}Obtaining semantic attributes which create interpretable relations between sequences $\nZ_i \in \nZC$ and the seen classes $\nYC$ (Fig.~\ref{fig:alg}).
	\textbf{2-}Using the obtained semantic attributes for efficient clustering of the unseen set $\nZC$.
\section{Multiple-Kernel Dictionary Learning Framework}
Similar to Fig.~\ref{fig:alg}, it is a common observation for real-world MTS data (e.g., human motions) to find partial similarities between different data classes when considering a subset of their dimensions.
Therefore, 
these similarities can lead to an interpretable description for a novel data sample (from $\nZC$) via its relation to the seen classes (from $\nYC$).
Furthermore, such a description leads to a better clustering of novel data points $\nZ_i$ without having any prior information on their class labels.
To achieve the above, 
we design a specific multiple-kernel dictionary (MKD) structure which is trained based on $\nYC$
and learns semantic attributes similar to Fig.~\ref{fig:alg}-left.
To be more specific, MKD combines dimensions of similar MTS samples in the feature space under non-negativity constraints.
These attributes can encode each unseen $\nZ_i \in \nZC$ as an interpretable description of its dimensions and to better separate it from previous (unknown) classes in $\nZC$ (Fig.~\ref{fig:alg}-right).

To be more specific,
we assume there exist $f$ non-linear implicit kernel functions $\{\Phi_i(\nY)\}_{i=1}^f$ to map each dimension of $\nY$ into an individual RKH-spaces 
\cite{jian2011design}.
A weighted combination of these kernels with individual coefficients $\beta_i\ge 0$ (entries of $\vec{\beta}$) induces an embedding of the data in the feature space as
$
\Phi(\nY,\nAlf):=
[
\sqrt{\nalf_{1}} \Phi_1(\nY)^\top \cdots \sqrt{\nalf_{f}} \Phi_{f}(\nY)^\top]^\top
$. 
We can apply this embedding to the whole training data via 
$
\Phi(\nYC,\nAlf):=[\Phi(\nY_1,\nAlf)\cdots\Phi(\nY_N,\nAlf)]
$,
and additionally we consider $k$ different weighting schemes of the individual kernels as $\nALF=[\nAlf_1 \cdots \nAlf_k] \in \mathbb{R}^{f\times k}$
to complement different existing classes in the data.
Now, We define our novel multiple kernel dictionary (MKD) matrix $\phiba$ as 
$$
\phiba:=[\Phi(\nYC,\nAlf_1)\na_1 \cdots \Phi(\nYC,\nAlf_k)\na_k] 
\qquad\mathrm{where\ }
\nA=[\na_1 \dots \na_k] \in \MBB{R}^{N \times k}
.
$$
Each dictionary column $\Phi(\nYC,\nAlf_i)\na_i$ is a weighted combination of selected dimensions and selected samples from $\nYC$ based on the value of $\vec{\beta}_i$ and $\na_i$ respectively. 
Due to the relation of $\Phi(\nYC,\nAlf_i)\na_i$ to different dimensions of $\nYC$, its columns can learn semantic attributes similar to those of Fig.~\ref{fig:alg}. 

To fit $(\nA,\nALF)$ to the data efficiently, we aim for the sparse reconstruction $\Phi(\nYC)\approx \phiba\nX$ in the feature space
based on a sparse matrix of codings $\nX=[\nx_1 \dots \nx_N] \in \MBB{R}^{k \times N}$.
%
%
%
To that aim, We propose the following  MKD sparse coding framework (MKD-SC) for training the dictionary parameters $(\nALF,\nA)$
and sparse codes $\nX$: 
\begin{equation}
\small
\begin{array}{ll}
\underset{\nALF,\nX,\nA}{\min}&
\|\Phi(\nYC)-\phiba\nX\|_F^2\\
\mathrm{s.t.} & \|\nx_i\|_0 < \nT,
~~\|\Phi(\nYC,\nAlf_i)\na_i\|_2^2=1
,~~\nas_{ij}, \nalf_{ij}, \nxs_{ij} \in \mathbb{R}^{+},~~\forall ij,
\end{array}
\label{eq:sp_feat}
\end{equation}
where 
$\nas_{ij}, \nalf_{ij}, \text{ and } \nxs_{ij}$ denote the $j$-th entry of the $i$-th column of $\nA,\nALF,\text{ and }\nX$ respectively. 
%
The loss term in Eq.~\ref{eq:sp_feat} measures the reconstruction error of the sparse coding based on the Frobenius norm $\|.\|_F$. 
The term $\|.\|_0$ denotes the $l_0$-norm
which employs sparsity constraints for elements of $\nX$ 
via the constant $\nT$ which results in having each $\nY_i$ constructed with sparse contributions from $\nYC$. 
The $l_2$-norm constraint on $\Phi(\nYC,\nAlf_i)\na_i$ prevents the optimization solutions from becoming degenerated \cite{rubinstein2008efficient}.

%
Hence the dictionary $\phiba$, which results from the optimization problem in Eq.~\ref{eq:sp_feat}, contains attributes (columns), which are weighted combinations of different exemplars and dimensions from $\nYC$. The non-negativity constraints result in
having similar resources become combined which leads to learning semantic attributes for $\phiba$ and an interpretable sparse description based on each $\nx_i$ \cite{hosseini2016non}.
In the Sec.~\ref{sec:partial} and ~\ref{sec:inc}, we benefit from this framework to describe and categorize unseen MTS samples.
\subsection{Optimization Scheme}\label{sec:optim}
We optimize the parameters $\nA$, $\nX$, and $\nALF$ in alternating steps, such that 
at each update step, we optimize Eq.~\ref{eq:sp_feat} with respect to one parameter while fixing the others.
Based on the dot-product relations
$\{\K_i(\nYC,\nYC)=\Phi_i(\nYC)^\top \Phi_i(\nYC)\}_{i=1}^f$,
it is possible to rewrite Eq.~\ref{eq:sp_feat} in terms of each of $(\nx_i, \na_i,\nAlf_i)$ individually to obtain a 
%
%
general convex form of
\begin{equation}
\begin{array}{l}
\underset{\ny}{\min}  ~\frac{1}{2}\ny^\top \MB{H} \ny+\vec{c}^\top \ny
\qquad \mathrm{s.t.}  ~\|\ny\|_0 < \nT , ~~ x_{i} \in \mathbb{R}^{+}~~\forall i,
\end{array}
\label{eq:quad}
\end{equation} 
in which $(\MB{H},\vec{c})$ are computed without any explicit reference to the embeddings $\Phi_i$. 
Such problems can be optimized via the non-negative quadratic pursuit (NQP\footnote{https://github.com/bab-git/NQP}) algorithm from \cite{hosseini2018confidentjmlr}.
Due to the page limit, we will put the detail regarding the reformulation of Eq.~\ref{eq:sp_feat} and the optimization steps in the online extended version of the paper \footnote{https://github.com/bab-git/MKD$\_$Unseen$\_$MTS}.
\subsection{Partial Reconstruction of Unseen MTS}\label{sec:partial}
In realistic MTS datasets such as human actions, it is expected to observe partial similarities between the dimensions of different classes. 
Therefore, we define the following error measure for the reconstruction of a selected set of dimensions $\MC{S}$ related to data $\nZ$: 
\begin{equation}
\nobj_{rec}^{\MC{S}}(\nZ,\nALF,\nA)=\|\MB{I}^{\MC{S}}\Phi(\nZ)-\MB{I}^{\MC{S}}\Phi_{\nALF}(\nA)\nX\|_2^2
/ \|\MB{I}^{\MC{S}}\Phi(\nZ)\|_2^2
\label{eq:partial}
\end{equation}
where $\nALF^{\MC{S}}$, and $\MB{I}^{\MC{S}}$ are modified versions of $\nALF$ and the identity matrix respectively via making all the entries zero except the rows corresponding to ${\MC{S}}$.
Consequently, the learned dictionary $\phiba$ can partially reconstruct the unseen time-series $\nZ$ for the subset $\MC{S}$ of its dimensions, if $\nobj_{rec}^{\MC{S}}(\nZ,\nALF,\nA)$ is relatively small.
\begin{algorithm}[!t]
	\caption{Incremental Clustering of an Encoded MTS data}
	\label{alg:inc}
	\KwInput{$\MB{R}$: Encoding of the new unseen data $\nZ$, 
		$\cal H$: The current hierarchical tree.}	
	\KwOutput{Place of $\nZ$ in the hierarchy $\cal H$.} 
	\textbf{If }$\exists C_n$ such that $d(\nZ,C_n) \leq \bar{d}(C_n)$ \textbf{then}\\ \label{line:sim}
	\qquad	\textbf{If }$C_n$ is a leaf node \textbf{then}
	add $\nZ$ to $C_n$\; \label{line:leaf}
	\qquad\qquad		\textbf{If }$({\bar{d}(C_{n1})+\bar{d}(C_{n2})})/{2\bar{d}(C_{n})}\le k_{clust}$ \textbf{then}\\ \label{line:split}
	\qquad\qquad\qquad		split $C_n$ into $C_{n1}$ and $C_{n2}$ using $k$-means\;
	\qquad\qquad\qquad				\textbf{If }$({\bar{d}(C_{n1})+\bar{d}(C_{n2})})/{2\bar{d}(C_{n})}\le k_{rmv}$ \textbf{then}\\ \label{line:replace}
	\qquad\qquad\qquad\qquad					Replace $C_n$ with $C_{n1}$ and $C_{n2}$\; 
	\qquad\qquad\qquad		\textbf{else }add $\{C_{n1},C_{n2}\}$ as the children of $C_n$\;
	\qquad	\textbf{else }Create a new child for $C_n$ as $C_{n_t}$ and add $z$ to it\;
	\textbf{else }Create a new leaf at the top level containing $\nZ$\;
\end{algorithm}
\subsection{Incremental Clustering of Unseen MTS}\label{sec:inc}
We propose Algorithm \ref{alg:inc} 
relying on the partial similarity of different MTS classes and the descriptive quality of the learned attributes of MKD.
This algorithm incrementally clusters the unseen sequences of $\nZC$ into a dendrogram $\MC{H}$ in an online fashion, and also finds the potential sub-clusters among them. 
To that aim, for each unknown MTS sequence $\nZ$, we prepare an encoding matrix $\MB{R} \in \MC{R}^{N \times f}$, $i$-th column of which represents the weights of contribution from $\nY$ in the reconstruction of the $i$-th dimension of $\nZ$. Therefore, 
$r_{ji}=\sum_{t=1}^{k}\nalf_{it}\nas_{jt}\nxs_t$
where $r_{ji}$ denotes the $j$-th entry of the $i$-th column of $\MB{R}$.
This matrix is considered as a rich encoded descriptor for dimensions of $\nZ$ based on $\nY$ and is used in Algorithm~\ref{alg:inc} to compare $\nZ$ to the previously categorized unseen data in $\MC{H}$ to find the best place for $\nZ$ in the dendrogram.
Line \ref{line:sim} of the algorithm finds $C_n$ as the most similar node to $\nZ$ based on the distance term
$d(\nZ,C_n)=\| \MB{R}_\nZ -{\overline{\MB{R}}}_{C_n}\|^2_F$,
and the intra-cluster distance for each node $C_n$ as 
$\bar{d}(C_n)=E_{\nZ_i \in C_{n}}[d(\MB{R}_{\nZ_i},{\overline{\MB{R}}}_{C_{n}})]$, 
where ${\overline{\MB{R}}}_{C_n}=E_{\nZ_i\in C_n}[{\MB{R}}_{\nZ_i}]$.
Regarding line~\ref{line:replace}, We choose $k_{rmv}=0.3$ in our experiments which results in an acceptable clustering outcome. 
\section{Experiments}\label{sec_exp}
To evaluate the performance of our sparse coding framework for representation and discrimination of unseen data, we choose the MTS datasets 
Cricket Umpire,
CMU mocap,
Articulatory Words,
and Squat with the descriptions provided by \cite{hosseini2016non}. 
For all the datasets, the Gaussian kernel matrices are computed as 
$\{\K_l(\nY_i,\nY_j)=exp(-\mathcal{D}_l(\nY_i,\nY_j)/\delta_l)\}_{l=1}^f$, 
where $\mathcal{D}_l(\nY_i,\nY_j)$ is the computed pairwise DTW-distance between the $l$-th dimension of $\nY_i$ and $\nY_j$ \cite{hosseini2016non} (but can be substituted with any other preferred distance). 
For tuning $\nT$ and the dictionary size in Eq.~\ref{eq:sp_feat}, we use 5-fold cross-validation.
\subsection {Partial Reconstruction Results}
In order to evaluate the reconstruction quality for each unseen data $\nZ$, we define the  
dimension-reconstruction accuracy measure as 
$DRA:=\frac{\# \text{ dimensions that } \{\nobj_{rec}^i(\nZ,\nALF,\nA) \leq 0.1\}}{\#\text{ total dimensions }}$
using Eq.~\ref{eq:partial}.
\begin{table}[!t]
	\caption{Average of DRA measure (\%) for reconstruction of the unseen classes.}
	\label{tab:part}	
	\centering
	\resizebox{0.5\textwidth}{!}{%
		\begin{tabular}{|l|c|c|c|c|} 
			\hline
			& Cricket & CMU & Words& Squat\\
			\hline
			DRA (\%)       &76.4&84.5&80.2&62.6\\
			\hline
		\end{tabular}
	}
\end{table}
\begin{figure}[b]
	\centering		
	\begin{subfigure}{0.4\textwidth}
		\centering		
		\includegraphics[width=1\linewidth]{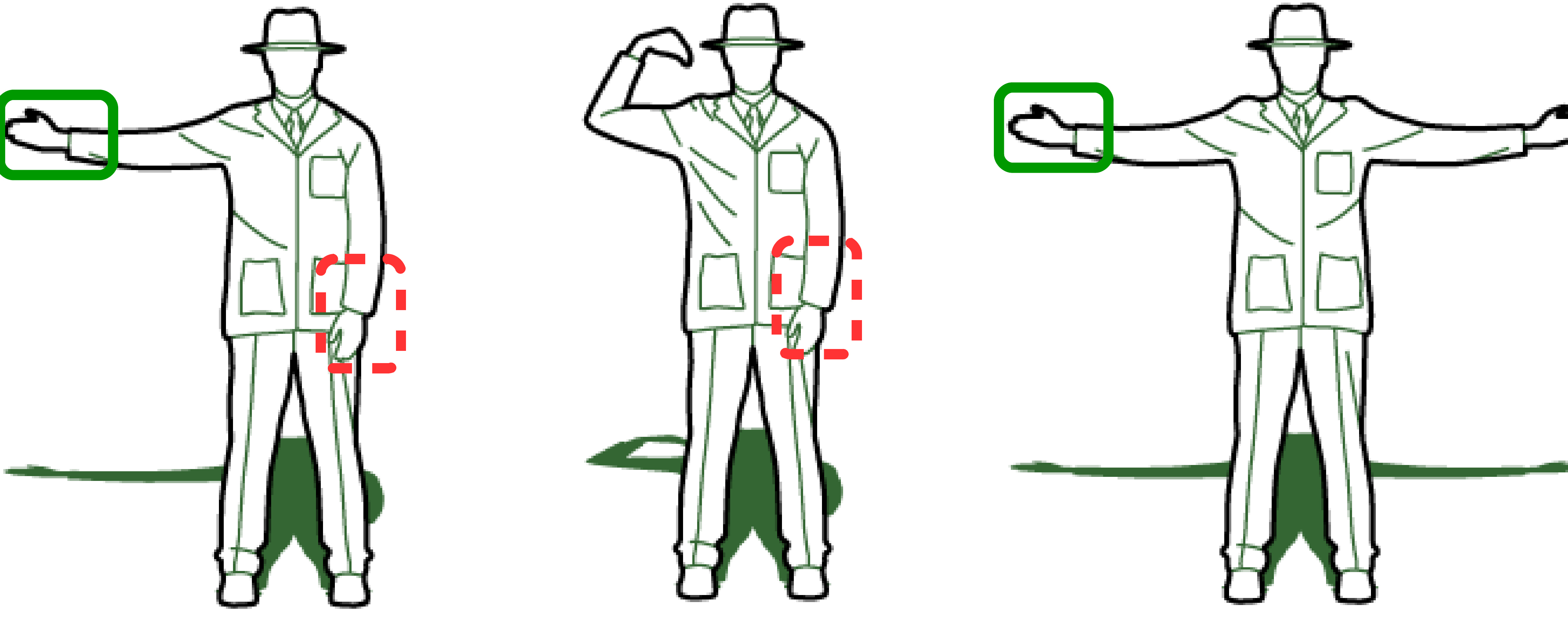}
		\caption{$no~ball~\rightarrow \{short+~~wide\}$}
	\end{subfigure}	
	\hspace{2cm}
	\begin{subfigure}{0.18\textwidth}
		\centering		
		\includegraphics[width=1\linewidth]{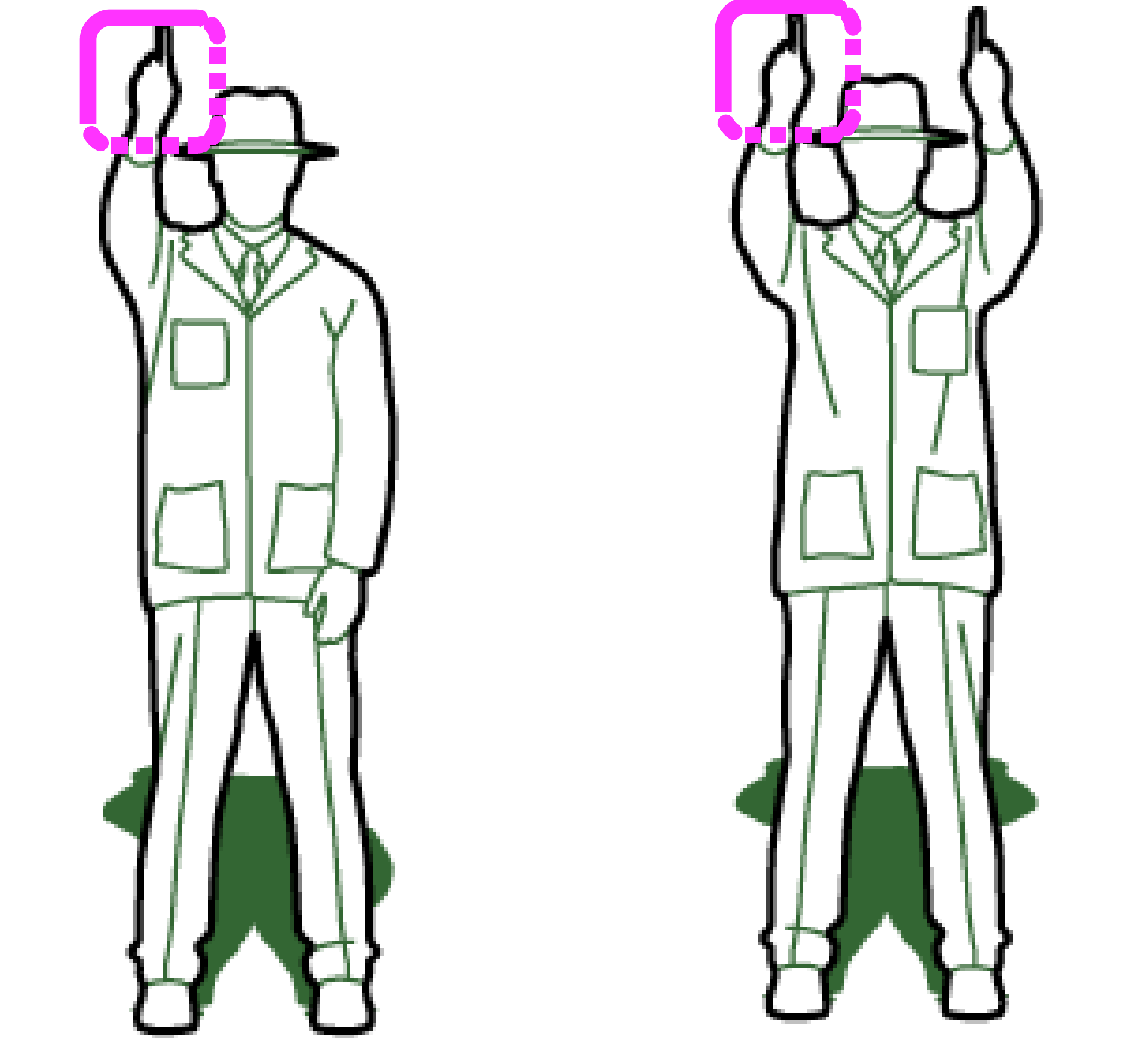}
		\caption{$out\rightarrow \{six\}$}		
	\end{subfigure}		
	\caption{Dimension-level interpretation of \textit{no-ball} and \textit{out} (Cricket) based on the training classes. Related dimensions are specified via using same-color rectangles.}
	\label{fig:part}
\end{figure}
Furthermore, each reconstructed dimension of $\nZ$ which satisfies the above threshold is interpreted via the class of data with the most contribution as in Sec.~\ref{sec:partial}.
Table~\ref{tab:part} reports the DRA values for the selected MTS datasets, 
where the CMU and Words datasets have higher DRA values due to their diverse set of training classes which increases the dimension-level similarity between seen and unseen classes.
As an example, We illustrate the dimension-level reconstruction of 2 unseen categories from the Cricket dataset in Fig.~\ref{fig:part}, 
in which the \textit{No ball} class is fully reconstructed via its relation to 
the movement of the left hand in the \textit{Short} class and to that of the right hand in the \textit{Wide} class. 
\subsection {Incremental Clustering Results}
To evaluate the incremental clustering of Sec.~\ref{sec:inc} 
we use the average clustering error (CE) and normalized mutual information (NMI) \cite{zhu2018nonlinear}. 
As the most relevant baseline, we choose the self-learning algorithm \cite{lu2016self} without its novelty detection part. 
Besides, 
we implement the spectral clustering algorithm on the original kernel matrix $\K(\nZ,\nYC)$ to compare our framework to the regular clustering of $\nZC$.
As another baseline, we also use the NNKSC algorithm \cite{hosseini2016non} as the single-kernel predecessor of MKD-SC, for which the $\MB{R}$ matrix becomes an $N$-dimensional vector.

\begin{table}[t]
	\caption{Clustering error (CE) (\%) and NMI the unseen categories.}
	\label{tab:clust}	
	\centering
	\resizebox{0.8\textwidth}{!}{%
		\begin{tabular}{|l||c|c||c|c||c|c||c|c|} 
			\hline
			\multirow{2}{*}{Methods} & \multicolumn{2}{c||}{Words} &  \multicolumn{2}{c||}{Squat}& 
			\multicolumn{2}{c||}{CMU} & \multicolumn{2}{c|}{Cricket}\\
			\cline{2-9}
			& CE & NMI & CE & NMI & CE & NMI & CE & NMI \\
			\hline
			{MKD-SC\scriptsize (Proposed)}       &12.31&0.89&0    &1   &9.28 &0.92&0&1\\
			\hline
			Self-learning \cite{lu2016self}      &18.75&0.84&0	  &1   &14.25&0.87&16.63&0.85\\
			\hline
			NNKSC\cite{hosseini2016non}  		 &21.61&0.78&15.74&0.88&18.88&0.85&12.45&0.87\\
			\hline												
			{Spectral Clustering}  							 &27.51&0.76&13.04&0.90&23.45&0.76&8.04&0.89\\
			\hline
		\end{tabular}}
	\end{table}
According to the clustering results in Table~\ref{tab:clust}, the proposed MKD-SC method provides encodings which lead to better clustering of the unseen data compared to the baselines.
The superiority of the spectral-clustering over NNKSC and self-learning methods (e.g., for Cricket dataset) depends on the discriminative quality of the original kernels.
Self-learning method can have a better performance than NNKSC and spectral-clustering when its descriptor-based features can better discriminate between the different categories of the unseen classes.
\section{Conclusion}\label{sec:conc}
In this research, we proposed an unsupervised framework which provides interpretable analysis of unseen classes in MTS datasets.
It is constructed based on a novel MKD structure which uses the kernel representations of MTS dimensions to learn semantic attributes.
%
%
%
%
%
Based on these attributes, 
our unsupervised MKD-SC framework reconstructs the unseen classes (partially/entirely) in the feature space according to the relation of their dimensions to those of the seen categories which provides an interpretable description of the novel data.
Based on the obtained sparse encodings, we proposed an incremental clustering to categorize novel MTS into distinct clusters gradually.
Experiments on real MTS benchmarks show the effectiveness of our MKD-SC framework in obtaining interpretable descriptions for unseen MTS classes. 
Additionally, the incremental clustering provides better clustering accuracy comparing to the baselines.
\section*{Acknowledgement}
This research was supported by the Cluster of Excellence Cognitive 
Interaction Technology 'CITEC' (EXC 277) at Bielefeld University, which
is funded by the German Research Foundation (DFG).
\bibliographystyle{unsrt}
\bibliography{/vol/semanticma/Thesis/Publications/Ref4Papers_CS}



\end{document}